\documentclass[runningheads]{llncs}

\usepackage[T1]{fontenc}
\usepackage{multirow}
\usepackage{subcaption}
\usepackage{booktabs}
\usepackage{amsmath}

\usepackage{tcolorbox}
\usepackage{xcolor, colortbl}

\usepackage{graphicx}
\usepackage{orcidlink}
\usepackage{hyperref}

\newif\ifblind
\blindfalse  % Set to \blindfalse for the final version

\begin{document}
\title{ComicsPAP: understanding comic strips by picking the correct panel}
%
%\titlerunning{Abbreviated paper title}
% If the paper title is too long for the running head, you can set
% an abbreviated paper title here
%
\ifblind
    \author{Anonymous Author(s)}
    \institute{}
    \authorrunning{A. Author(s) et al.}
\else
    \author{Emanuele Vivoli\thanks{contributed equally}\inst{1,2}\orcidlink{0000-0002-9971-8738} \and
    Artemis Llabrés$^{\tiny\star}$\inst{1}\orcidlink{0000-0002-6128-1796} \and
    Mohamed Ali Souibgui\inst{1}\orcidlink{0000-0003-0100-9392} \and\\
    Marco Bertini\inst{2}\orcidlink{0000-0002-1364-218X} \and
    Ernest Valveny Llobet\inst{1}\orcidlink{0000-0002-0368-9697} \and
    Dimosthenis Karatzas\inst{1}\orcidlink{0000-0001-8762-4454}}

    \authorrunning{E. Vivoli, A. Llabrés et al.}
    % First names are abbreviated in the running head.
    % If there are more than two authors, 'et al.' is used.
    %
    \institute{CVC, Autonomous University of Barcelona, Spain\and
    MICC, University of Florence, Italy\\
    \email{\{evivoli, allabres\}@cvc.uab.cat}}
    
  % Use proper running heads here if needed
\fi

\maketitle              % typeset the header of the contribution

\begin{figure}
\vspace{-10mm}
\hspace{-20mm}
    \begin{subfigure}{0.7\linewidth}
        \centering
        \includegraphics[height=0.32\textheight]{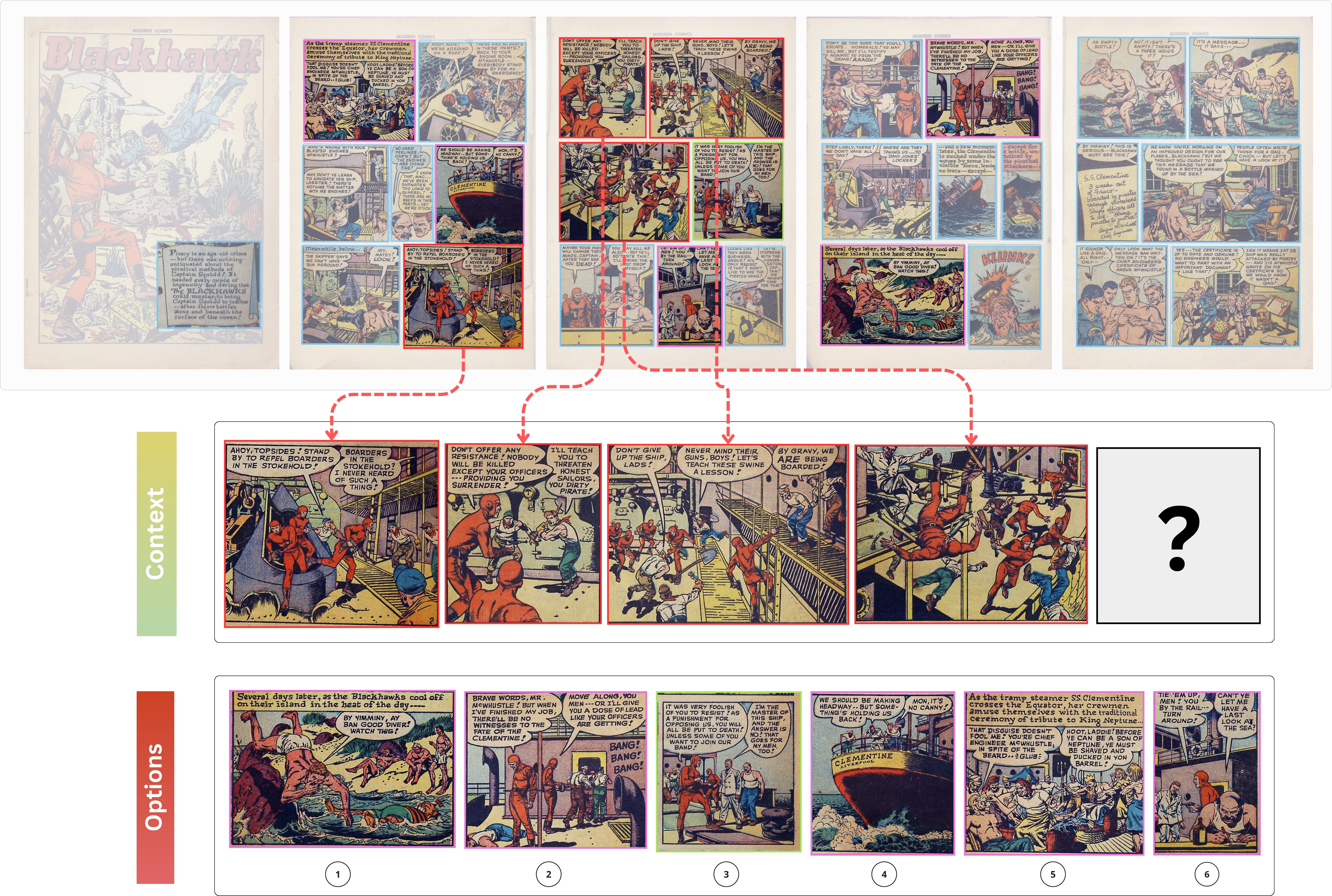}
    \end{subfigure}
    \hspace{15mm}
    \begin{subfigure}{0.25\linewidth}
        \centering
        \includegraphics[height=0.32\textheight]{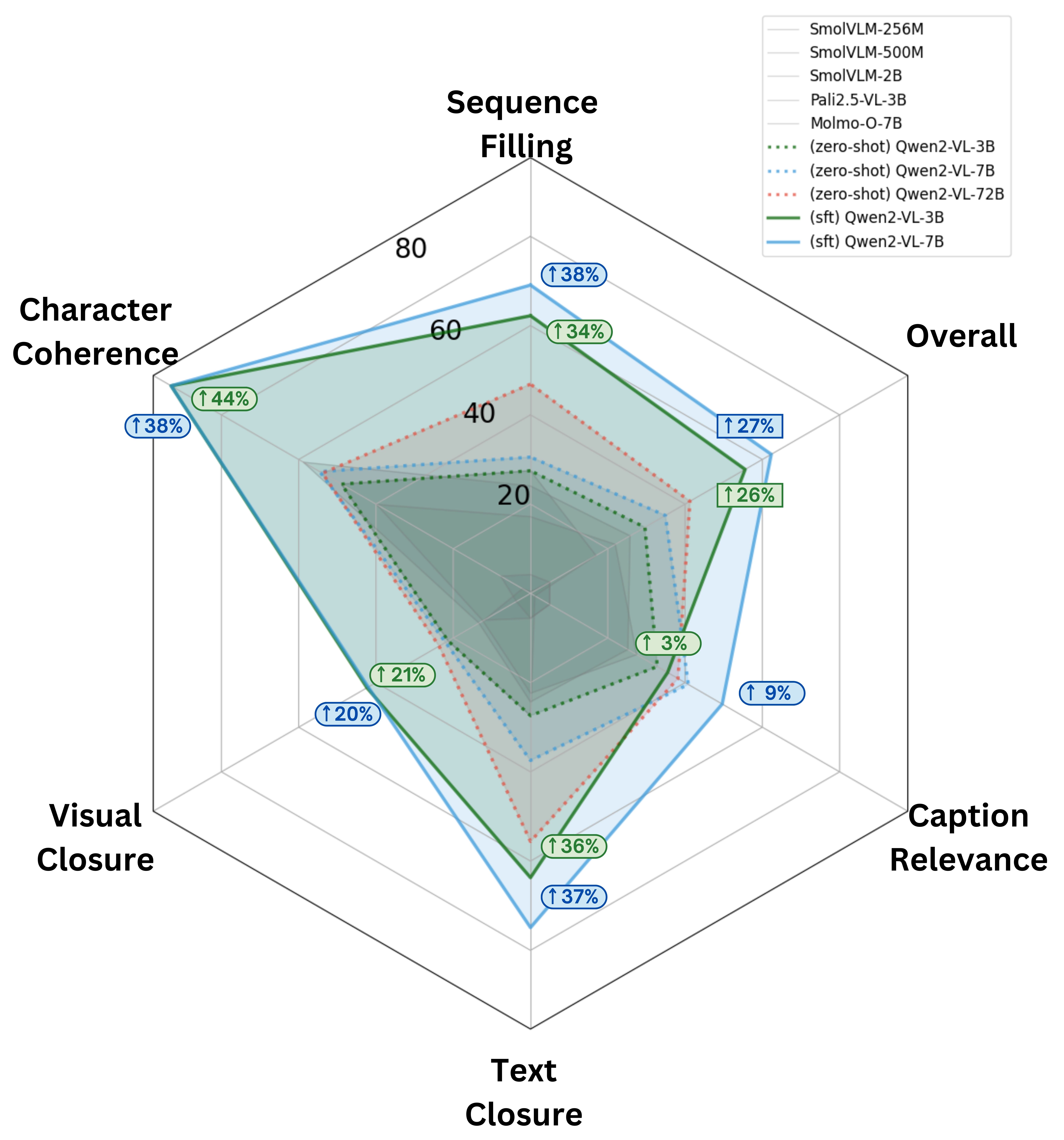}
    \end{subfigure}
    \caption{ComicsPAP dataset composition (left) and our Comic-adapted LMM performances vs. zero-shot LMMs (right).}
    \label{fig:teaser}
\end{figure}
\vspace{-10mm}
\begin{abstract}
Large multimodal models (LMMs) have made impressive strides in image captioning, VQA, and video comprehension, yet they still struggle with the intricate temporal and spatial cues found in comics. To address this gap, we introduce \textit{ComicsPAP}, a large-scale benchmark designed for comic strip understanding. Comprising over 100k samples and organized into 5 subtasks under a \textit{Pick-a-Panel} framework, \textit{ComicsPAP} demands models to identify the missing panel in a sequence. Our evaluations, conducted under both multi-image and single-image protocols, reveal that current state-of-the-art LMMs perform near chance on these tasks, underscoring significant limitations in capturing sequential and contextual dependencies. To close the gap, we adapted LMMs for comic strip understanding, obtaining better results on \textit{ComicsPAP} than 10x bigger models, demonstrating that \textit{ComicsPAP} offers a robust resource to drive future research in multimodal comic comprehension.

\end{abstract}
\section{Introduction}
% - LMMs are great at documents, video, images, etc.
% - comics is built different from other medium
% - many tasks on comics have been proposed and tackle, but we don’t know if LMMs understand comic strips yet
% - so we propose ComicsPAP, we benchmarks LMMs for comics strips understanding, and adapt LMMs for this
% - we hope our datasets and models to give the comunity a sense of what these models are capable of (because they are often limited)

% why comics
Large multimodal models (LMMs) have achieved remarkable progress in understanding diverse data modalities, excelling at tasks such as image captioning~\cite{deitke2024molmopixmoopenweights,bai2025qwen25vltechnicalreport}, visual question answering (VQA)~\cite{bai2025qwen25vltechnicalreport,chen2024fargpt4vclosinggap,zhang2025videollama3frontiermultimodal,deitke2024molmopixmoopenweights}, and video comprehension~\cite{tang2024videounderstandinglargelanguage}. These success stories share a crucial advantage: the underlying data are naturally segmented (e.g., video frames) and exhibit strong temporal or spatial coherence, easing the burden of cross-frame reasoning.
Comic strips, by contrast, deliberately disrupt these assumptions. Each panel depicts a distinct moment, producing abrupt visual shifts that break the smooth continuity exploited by video models. Furthermore, segmentation is implicit—readers infer panel boundaries from gutters whose presence, shape, and clarity are highly variable. Linguistic signals are likewise fragmented: dialogue and narration appear in localized bubbles or captions and must be stitched together across non-adjacent panels. Effective comic comprehension therefore hinges on three intertwined capabilities: (i) inferring latent panel boundaries, (ii) resolving recurring characters under diverse artistic styles, and (iii) reconstructing causal story lines from sparse, discontinuous multimodal evidence.
These challenges position comics as a rigorous stress test for next-generation vision–language models.

% what existed already?
Recent studies have begun to explore the intricacies of comics understanding. A comprehensive survey~\cite{vivoli2025missingpiecevisionlanguage} has highlighted both established and emerging tasks aimed at modeling and understanding comics using machine learning techniques. 
% While evaluating the capabilities of state-of-the-art vision-language models in this domain, 
While advanced tasks such as comic story comprehension and multi-page comic generation have been proposed, even more fundamental challenges—such as single-page understanding~\cite{sachdeva_manga_2024,sachdeva_tails_2024,vivoli_comix_2024} and single-panel generation~\cite{wu2024diffsenseibridgingmultimodalllms}—remain challenging for state-of-the-art vision-language models. Despite progress in object detection~\cite{vivoli_comics_2024,sachdeva_manga_2024}, full-page analysis~\cite{vivoli_comix_2024}, and single-panel interpretation~\cite{vivoli2024comicapvlmspipelinedense}, it is still unclear whether foundational models can comprehend the sequential narrative across multiple panels.

% what we propose
In this work, we address the lack of comic strip analysis by benchmarking foundational models for comic panel sequence understanding. Our contributions are threefold: (i) we introduce the \textit{ComicsPAP} dataset, structured around three distinct skills and five subtasks, which reframes sequence understanding as the task of selecting the appropriate panel that best fits within a given ordered set—an approach we term ``Pick A Panel''; (ii) we evaluate state-of-the-art LMMs on \textit{ComicsPAP} and demonstrate that their performance is no better than random choice; (iii) we fine-tune models with 3B and 7B parameters on \textit{ComicsPAP}, achieving superior performance compared to larger models on this task.

\ifblind
All the material, the \textit{ComicsPAP} dataset, the fine-tuned model weights, as well as the training and evaluation code, will be publicly available.
\else
The \textit{ComicsPAP} dataset is publicly available on HuggingFace\footnote{\url{https://huggingface.co/datasets/VLR-CVC/ComicsPAP}}, as are the fine-tuned model weights\footnote{\url{https://huggingface.co/VLR-CVC/Qwen2.5-VL-3B-Instruct-lora-ComicsPAP}}. The training and evaluation code are available in the GitHub repository\footnote{\url{https://github.com/llabres/ComicsPAP}}, and an evaluation server is hosted on the competition page\footnote{\url{https://rrc.cvc.uab.es/?ch=31\&com=tasks}}.
\fi

\section{Related works}

\paragraph{\textbf{Data Availability.}} Availability and copyrighted data remain the foremost challenges in comics analysis. Many previous studies have addressed these issues by linking to original sources to collect images while only sharing the accompanying annotations~\cite{vivoli_comics_2024,vivoli_comix_2024}. However, such approaches face significant limitations when scaled up. For instance, source material may become inaccessible over time, resulting in non-reproducible experimental settings (over 50\% of LAION images~\cite{schuhmann2021laion400mopendatasetclipfiltered} are no longer available). An alternative strategy involves providing scripts to download images directly from the original sources~\cite{sachdeva_manga_2024} and sharing only cropped versions~\cite{sachdeva_tails_2024}. Yet these methods also have drawbacks: sources can expire, and users might lack the necessary rights to download copyrighted images. To overcome these challenges, we exclusively utilize non-copyrighted, American-comic-styled materials from the renowned DCM website\footnote{\href{https://digitalcomicmuseum.com/index.php}{https://digitalcomicmuseum.com/index.php}}. This choice enables us to share both the images and their annotations freely.

\paragraph{\textbf{Panel Sequence Analysis in Comics.}} Early work in panel sequence analysis, such as COMICS~\cite{iyyer2017amazingmysteriesgutterdrawing}, introduced ``closure'' tasks where models select the correct subsequent panel from options based on visual cues, combined visual and textual features, or character-driven attributes. However, these methods relied on old-generation OCR and panel extraction, resulting in poor-quality annotations and distractors sampled from unrelated pages—issues that rendered the task relatively trivial (with recent transformer approaches yielding already significant improvements). Subsequent work like ComicVT5~\cite{vivoli_multimodal_2024} improved OCR accuracy, sampled distractor panels from adjacent pages, and extended the task to text generation rather than classification. More recently, MangaUB~\cite{ikuta2024mangaubmangaunderstandingbenchmark} extended these evaluations to manga panels with tasks ranging from single-panel analysis (predicting weather, time, and counting characters) to comic strip understanding, yet even basic inferences proved challenging for LMMs. Complementing these efforts, ComiCap~\cite{vivoli2024comicapvlmspipelinedense} showed that, with proper model selection, vision-language models can generate detailed captions in a zero-shot setting, underscoring the potential for further advances in comic panel comprehension.
\newline\\
% what is missing
Despite these advances, panel sequence analysis has largely stagnated with tasks based on earlier, automatically generated annotations for next-panel classification using randomly sampled alternatives. This approach suffers from several intrinsic limitations: (i) comics are seldom segmented into coherent stories before analysis, so randomly sampled panels often come from different narratives, styles, and character sets, making the task inherently easier; (ii) tasks are predominantly framed as next-panel prediction, neglecting the evaluation of causal reasoning through the reconstruction of sequences with missing panels; and (iii) the focus remains largely visual, with little emphasis on selecting the correct panel based solely on textual descriptions.

In this work, we build upon these foundations by creating a dataset designed around specific skill sets for comics comprehension. We systematically identify both existing and novel tasks that highlight these underlying capabilities. Moreover, we benchmark state-of-the-art open-weight LMMs on our \textit{ComicsPAP} dataset and provide a held-out test set and a validation set, thereby providing the comics and LMM communities with a robust platform for model evaluation.

\section{Dataset Creation}

This section describes the construction and annotation procedures of the \textit{ComicsPAP} dataset. Our approach is driven by explicitly defining the cognitive skills we aim to assess and designing corresponding tasks that rigorously evaluate these skills. Specifically, we focus on three primary multimodal understanding skills—\textit{narrative anticipation}, \textit{multi-frame reasoning}, and \textit{co-reference resolution}. 

\subsection{Tasks definition}
\begin{figure}[t]
\centering
\includegraphics[width=\linewidth]{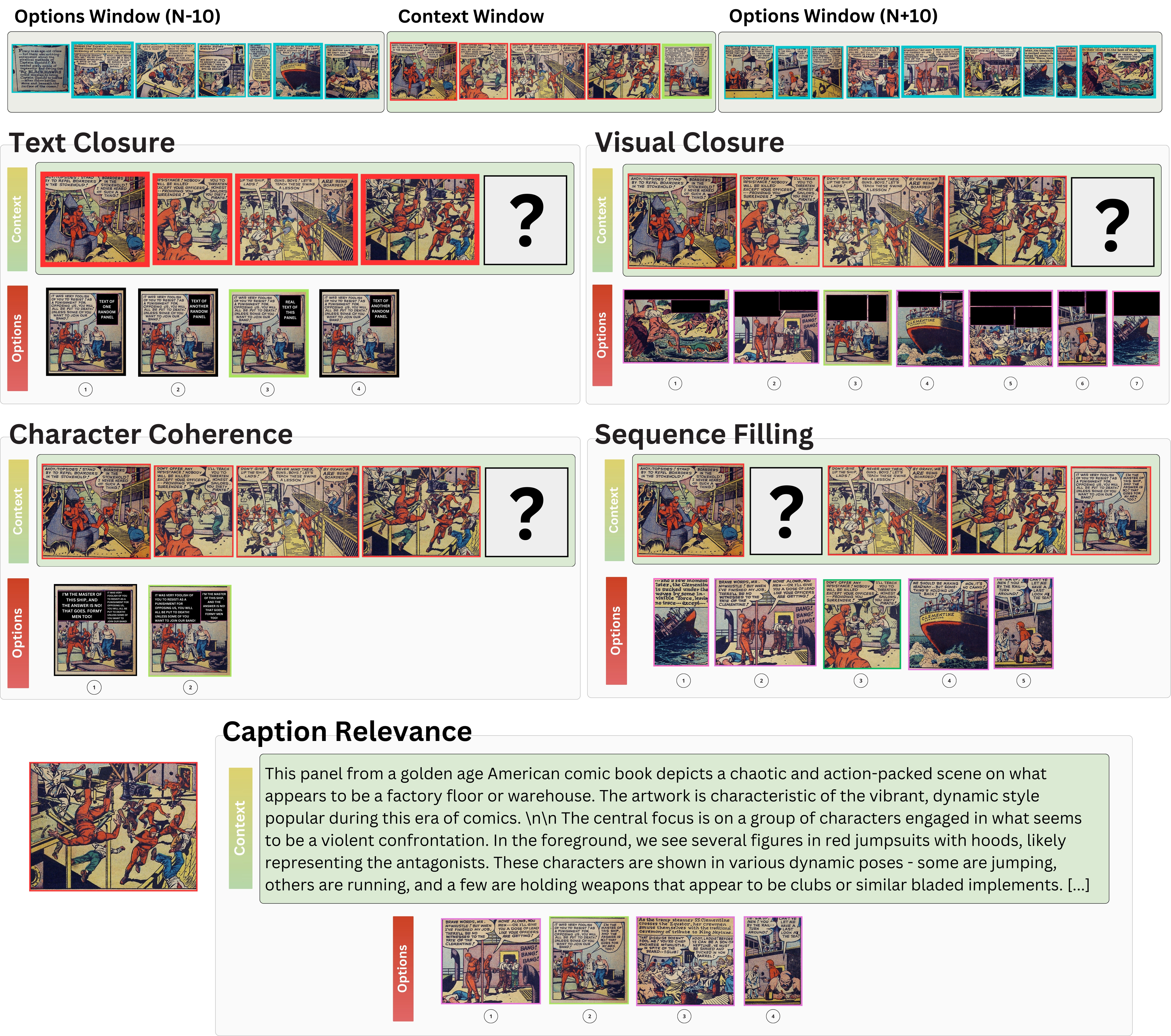}
\caption{Overview of the five tasks in \textit{ComicsPAP}. In \textit{Caption Relevance}, the image is shown only for illustrative purposes: the real task doesn't provide the image.}
\label{fig:tasks_overview}
\end{figure}

Each skill is operationalized through carefully designed tasks under a unified evaluation paradigm we term \textbf{``Pick A Panel''}, where models select the correct panel given multimodal context and options. Figure \ref{fig:tasks_overview} provides a visualization of the five tasks.

\paragraph{Narrative Anticipation (Closure Tasks).}
The narrative anticipation skill measures a model's capability to infer immediate narrative continuation (or next-panel), similar to the next-token prediction task in NLP~\cite{radford2018improving}. It relies on unidirectional contextual information derived from preceding panels, assessing the model’s understanding of narrative structure and visual-textual continuity.
% tasks origin
We adapt the well-established closure tasks from the COMICS dataset~\cite{iyyer2017amazingmysteriesgutterdrawing} into three refined variants: \textbf{character coherence}, \textbf{visual closure}, and \textbf{text closure}. In particular, \textit{visual closure} predicts the next panel using only visual features (hiding text in candidate panels), \textit{text closure} predicts the correct text in the next panel (showing the next panel with different options for the text), and \textit{character coherence} predicts the text associated to each character (showing the next panel with varying options of character-text linking). All these tasks provide three consecutive context panels, requiring models to select the correct subsequent (4th) panel among distractors sampled from the same coherent narrative segment. This addresses a key limitation from prior work, where distractors came from different books, allowing potential stylistic or character-based shortcuts.
% task
In the \textit{visual closure} variant, all context textboxes remain visible, and distractors (six total) are sampled within a challenging $\pm10$-panel window around the context panels, substantially increasing task difficulty compared to prior ``hard'' settings. The \textit{text closure} variant further standardizes evaluation by dynamically rendering texts within bounding boxes. To maintain difficulty and fairness, preventing models from exploiting visual font-size cues, distractors in \textit{text closure} tasks are pre-filtered by text length similarity with the correct one. Additionally, \textit{character coherence} specifically selects sequences where the final panel contains exactly two prominent textboxes. A textbox is assigned to a panel if at least 80\% of its area overlaps with the panel bounding box.

\paragraph{Multi-Frame Reasoning (Sequence Filling Task).}
In contrast to unidirectional closure tasks, multi-frame reasoning evaluates a model's ability to integrate both preceding and succeeding context, inspired by the masked language modeling strategy popularized by BERT~\cite{devlin2019bert} and inspired by the Cloze task \cite{taylor1953cloze}. This skill requires holistic narrative comprehension, including causal and temporal inference across multiple panels.
% task
We formalize this skill in the \textbf{sequence filling task}, employing a five-panel sliding window from comic narratives with one randomly masked panel. The model selects the correct missing panel from six candidate panels of the masked position, sampled within the same $\pm10$ panel window as previously indicated. This design ensures comprehensive contextual reasoning from both directions, directly testing models' capacity for narrative gap-filling without further task-specific filtering.

\paragraph{Co-reference Resolution (Caption Relevance Task).} 
Co-reference resolution involves correctly interpreting implicit references across comic panels, which means finding implicit references in consecutive panels.
% relations between consecutive panels
Following prior analysis~\cite{iyyer2017amazingmysteriesgutterdrawing,cohn_visual_2025}, we recognize four major intra-panel transitions in comics~\footnote{We anticipate similar proportions as our dataset is randomly sampled from a subset (Comics100) of the original COMICS collection.}: (i) \textit{action-to-action} ($34\%$), where the previous action continues in the next panel; (ii) \textit{subject-to-subject} ($32.7\%$), where the same subject appears with a closely related action; (iii) \textit{continued conversation} ($17.7\%$), where dialogue flows seamlessly between panels; and (iv) \textit{scene-to-scene} ($13.8\%$), where a complete change in setting occurs. The remaining panels, designated as moment-to-moment, represent the same exact moment, while less than $0.57\%$ are parallel transitions where both visual and textual content vary drastically.
% motivation
Based on these intra-panel transitions, we want the model to find the correct panel given some sort of information about the previous one, not favoring any of the previously described transition types. Thus, we use text description: a detailed caption of the previous panel.
% task
To evaluate this skill, we introduce the \textbf{caption relevance task}, where models select the subsequent panel corresponding to a given caption from among the next five sequential panels in the narrative. Captions are generated by a specialized model (detailed in later sections) to ensure consistency. To mitigate straightforward linguistic cues—particularly in continued conversations—we explicitly exclude spoken text from generated captions.
While prior work demonstrated success with explicit textual cues such as "the first page" or "the second image"~\cite{jiang2024mantisinterleavedmultiimageinstruction}, \textit{ComicsPAP} emphasizes implicit co-reference by requiring models to link detailed captions with adjacent panels. This skill expects models to infer coherent relationships across characters, settings, or semantic continuity, especially during scene transitions~\cite{iyyer2017amazingmysteriesgutterdrawing}.

Overall, these tasks are formulated as multi-class classification problems with text-panel inputs, and their design is aimed at probing different facets of comic comprehension in a cohesive manner.

\subsection{Benchmark Annotations}
We adopt the Comics100 collection from the Comics Dataset Framework~\cite{vivoli_comics_2024} and follow the prescribed test, validation, and train splits. Unlike previous methods that randomly partition pages~\cite{guerin2013ebdtheque,nguyen2018digital}, we split entire books to preserve narrative continuity. To rigorously evaluate the skills targeted by our tasks, we manually annotated the test and validation splits as follows.

\textit{\textbf{Stories Splitting.}} For the test and validation books, we performed manual Page Stream Segmentation~\cite{vivoli2025missingpiecevisionlanguage} to extract coherent stories composed of consecutive pages. In this process, non-content pages such as covers, first pages, advertisements, letters from authors, editors, or readers, as well as standalone text stories, were removed.

\textit{\textbf{Detection and Transcriptions.}} In parallel, panels and textboxes were manually detected and assigned a reading order. Initial transcriptions of the textboxes were automatically generated using Magi~\cite{sachdeva_manga_2024} and subsequently refined by human annotators.

\textit{\textbf{Captions.}} For caption generation, we leveraged the state-of-the-art Molmo-72B model\footnote{\href{https://huggingface.co/allenai/Molmo-72B-0924}{https://huggingface.co/allenai/Molmo-72B-0924}}~\cite{deitke2024molmopixmoopenweights}, based on the Qwen2-72B language model\footnote{\href{https://huggingface.co/Qwen/Qwen2-72B}{https://huggingface.co/Qwen/Qwen2-72B}}~\cite{yang2024qwen2technicalreport} and trained on the novel Pixmo dataset. The model was constrained to output up to 512 tokens per caption (with most captions being considerably shorter) and was prompted to exclude spoken or narration text.

\begin{figure}
    \centering
    \includegraphics[width=\linewidth]{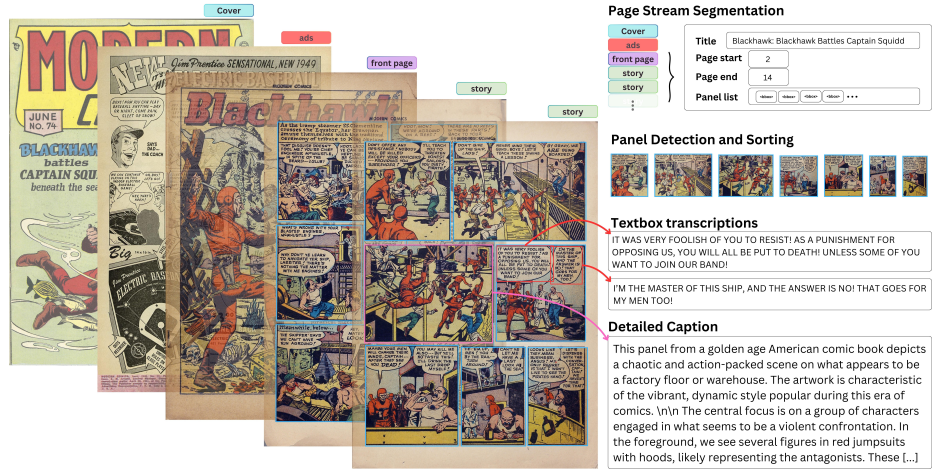}
    \caption{Overview of validation and test annotations.}
    \label{fig:annotations}
\end{figure}

\begin{figure}
    \centering
    \includegraphics[width=\linewidth]{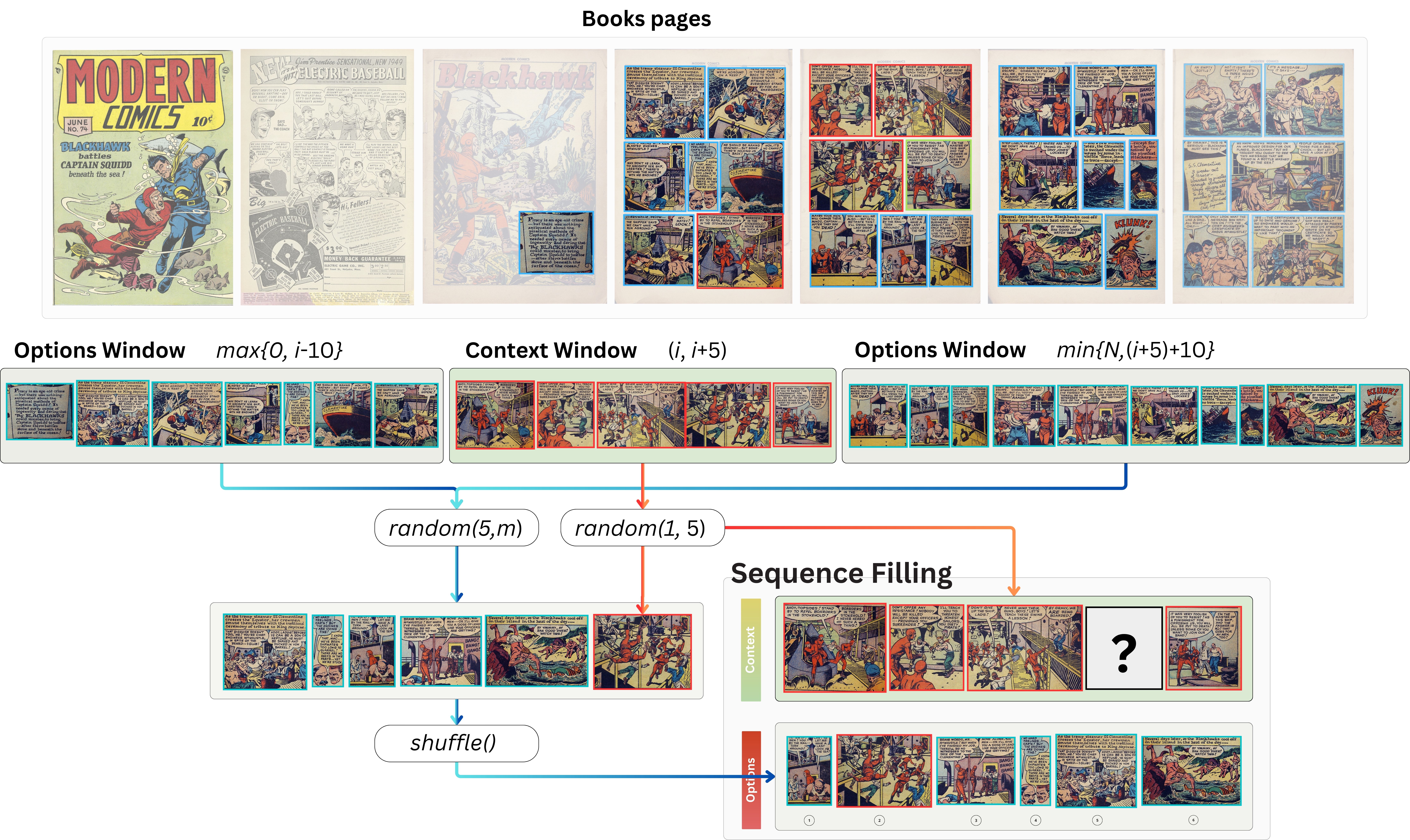}
    \caption{Overview of automatic task creation from the manually annotated story: the story pages (top); panel detection and ordering (middle); and the $N=5$ context and $M=6$ randomly sampled options (bottom).}
    \label{fig:sliding_window}
\end{figure}

Figure~\ref{fig:annotations} illustrates an overview of the annotation quality. With these elements in place—stories, panels, textbox coordinates, OCR transcriptions, panel reading orders, and panel captions—we automatically generated our five tasks via an ensemble of three components: (i) a context sliding window, (ii) random distractor sampling, and (iii) rendering. 
For all tasks, a sliding window of $N$ consecutive panels (based on the annotated order) is used, where $N=5$ for all tasks except \textit{caption relevance} ($N=1$). A larger sliding window spanning from $N-10$ to $N+10$ is employed to sample random distractors, from which the context panels are removed. The number of distractor options, $M$, varies by task: $M=6$ for \textit{visual closure}, $M=4$ for \textit{sequence filling}, $M=3$ for \textit{caption relevance} and for \textit{text closure} (sampling constrained by similar text length), and $M=0$ for \textit{character coherence}. 
This variability stems from task-specific design considerations inherited from the original closure tasks~\cite{iyyer2017amazingmysteriesgutterdrawing}, where \textit{text closure} initially had three distractors and \textit{character coherence} naturally offers only two possible text-swaps, making distractors unnecessary. For \textit{caption relevance}, we follow the distractor count of the original \textit{text closure} task with $M=3$. Inspired by recent efforts to increase task complexity~\cite{vivoli_multimodal_2024}, we deliberately increased distractor count for \textit{visual closure} ($M=6$). Similarly, we opted for a larger distractor pool in our newly introduced \textit{sequence filling} task, enhancing complexity beyond standard closure settings. An overview illustrating the quality of these annotations and task generation is provided in Figure~\ref{fig:annotations}.
An illustration of the sliding window and distractor sampling for visual closure is provided in Figure~\ref{fig:sliding_window}.

\subsection{Automatic Training Annotations}
Initial experiments revealed that state-of-the-art LMMs perform near chance on our benchmark tasks, motivating us to leverage a larger training set to fine-tune these models. For the remaining 80 training books, we employed automatic annotation techniques to scale dataset creation while preserving quality.

Although robust automated ``Panel Stream Segmentation'' remains challenging, we manually segmented the training books into coherent stories to ensure narrative consistency. For the remaining annotations, we utilized state-of-the-art models: FasterRCNN~\cite{vivoli_comics_2024} for panel and textbox detection, and Magi~\cite{sachdeva_manga_2024} for OCR. We modified the Magi algorithm to adhere to the comics’ left-to-right, top-to-bottom reading order by adjusting the construction of the Directed Acyclic Graph for child panels.

We employed the same Molmo-72B model for caption generation and applied the identical sliding window and random distractor sampling algorithm described above to create training samples. Notably, while non-overlapping sliding windows were used for the test and validation splits, we allowed overlapping sliding windows for the training set to maximize sample diversity from the 80 stories. This approach yields a rich training set, enabling effective fine-tuning of LMMs for improved comic strip understanding.

\subsection{Summary of Dataset Statistics}
Table~\ref{tab:dataset_sizes} summarizes the dataset sizes for each task across the training, validation, and test splits. These statistics underscore the scale and diversity of our dataset, which serves both as a rigorous evaluation benchmark (with more than 3k test samples) and as a valuable resource for fine-tuning LMMs (with over 99k unique samples).

\begin{table}[t]
\centering
\caption{Dataset sizes for each task across the training, validation, and test splits.}
\vspace{5mm}
\label{tab:dataset_sizes}
\begin{tabular}{lccc}
\toprule
\textbf{Task} & \textbf{Train} & \textbf{Val} & \textbf{Test} \\
\midrule
Sequence Filling     & 23,604  & 262   & 932   \\
Text Closure         & 17,898  & 259   & 924   \\
Visual Closure       & 24,166  & 300   & 1000  \\
Character Coherence  & 10,157  & 143   & 489   \\
Caption Relevance    & 23,604  & 262   & 932   \\
\midrule
\textbf{Total}       & 99,429  & 1,226 & 3,278 \\
\bottomrule
\end{tabular}
\end{table}

\section{Experiments and Benchmarks}

In this section, we describe our experimental setup and report benchmark results on the \textit{ComicsPAP} tasks. We first evaluate a range of large multimodal models (LMMs) in a zero-shot setting, covering both small and large parameter regimes. We then assess the impact of fine-tuning on our automatically generated training set.

To assess comic strip understanding, we evaluate a variety of LMMs spanning small and large parameter regimes. SmolVLM~\cite{marafioti2025smolvlm} in 256M, 500M, and 2.2B parameters. Two 3B parameter models: PaliGemma2~\cite{steiner2024paligemma2familyversatile}, and Qwen2.5-VL-3B~\cite{Qwen2VL}. For large models, we have considered the following 7 billion parameter models: Qwen2.5-VL-7B~\cite{Qwen2VL}, Molmo-7B~\cite{deitke2024molmopixmoopenweights} (both the -D and -O variants), and the larger 72B parameter Qwen2.5-VL-72B~\cite{Qwen2VL}.

\begin{figure}[t]
    \centering
    \includegraphics[width=\textwidth, height=0.3\textheight]{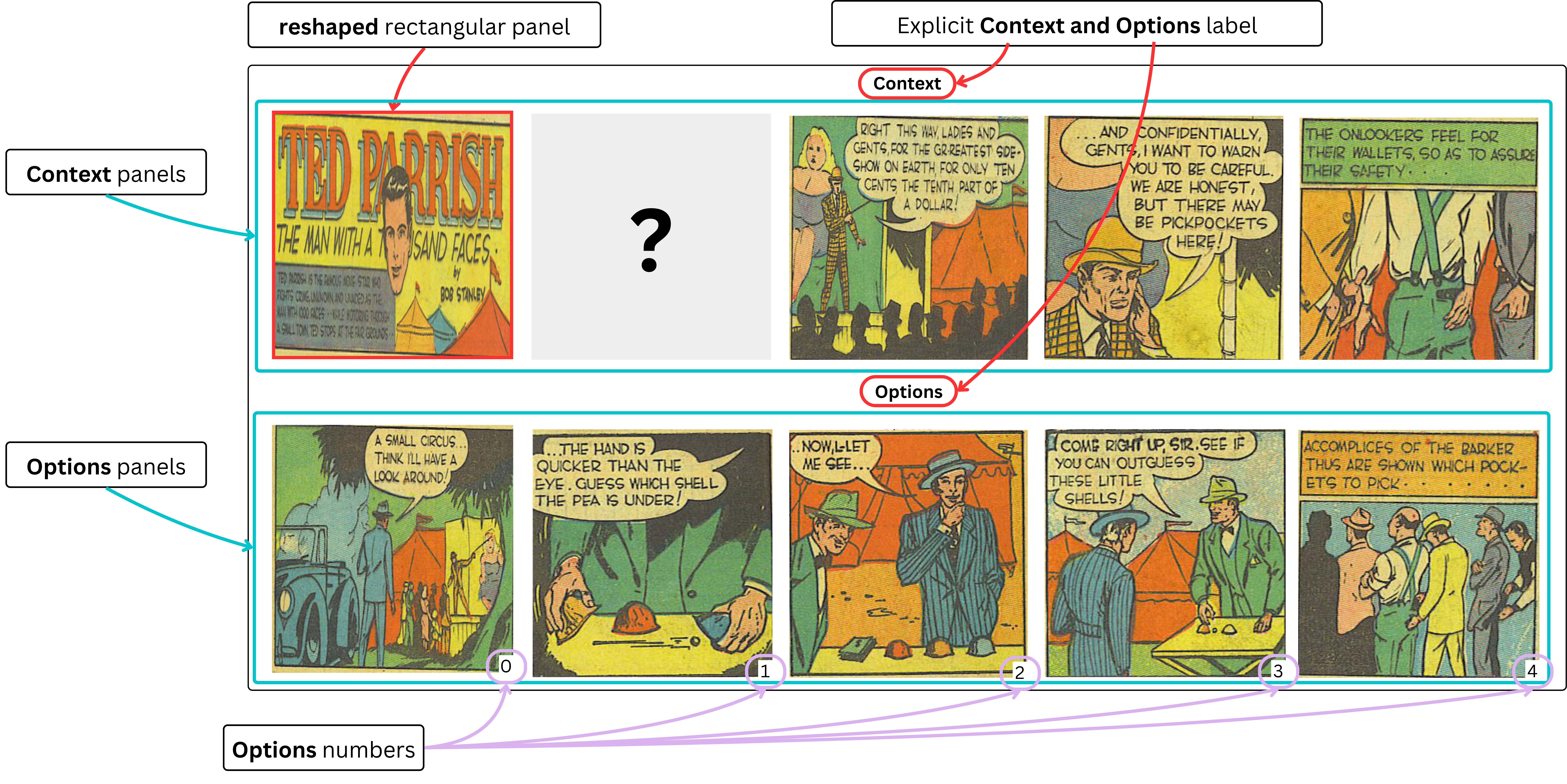}
    \caption{Single image example for the task of \textit{sequence filling}. All panels are reshaped as squared images; panels are located to minimize waste of space; and option numbers are provided at the bottom of the panels.}
    \label{fig:single_panel_example}
\end{figure}

Since the \textit{ComicsPAP} tasks involve multiple context and option panels, we must carefully consider how these panels are presented to the models. While many LMMs are capable of processing interleaved text and images, our experiments revealed that they often struggle to differentiate between context and option panels. To mitigate this, we render all panels as a single image. In all examples but caption relevance, the top row contains the context panels with a gray square (marked with a question mark) indicating the missing panel and the bottom row displays the candidate options (see Figure~\ref{fig:single_panel_example}). For the Caption Relevance task, the single image contains only the option panels, with the caption provided as additional text in the prompt. As an ablation study, we tested two scenarios—rendering the caption visually and providing it as text tokens—and found that visually rendering the caption resulted in slightly lower performance.

\subsection{Zero-shot Benchmarks}

For zero-shot evaluation, we designed a straightforward prompt that instructs the model to return an answer in the format ``\texttt{answer: <choice number>}''. An example prompt is shown in Figure~\ref{fig:prompt}. We adapt this prompt slightly for each model to account for their individual peculiarities.

\begin{figure}[t]
    \centering
    \begin{tcolorbox}[width=0.8\textwidth, colframe=black, colback=white]
    Pick A Panel Task: In the image you have two rows of comic panels. The top row is the context and the bottom row is the options. The context row has a missing panel marked with a question mark. Choose the option that best fits the missing panel. You must return your final answer as a number with ``\texttt{answer: <choice number>}''
    \end{tcolorbox}
    \caption{Prompt for zero-shot evaluation on the ComicPAP dataset.}
    \label{fig:prompt}
\end{figure}

Table~\ref{tab:zero-shot} reports the per-task and overall global accuracy (in \%) for the evaluated models. Global accuracy is computed as the weighted average over the five tasks: Sequence Filling, Character Coherence, Text Closure, Visual Closure, and Caption Relevance. In our zero-shot experiments, most models—regardless of size—produce outputs close to random chance. In particular, SmolVLM models yield especially low accuracies, as they often fail to interpret the prompt and instead describe the panels. Although models such as PaliGemma2 and the Molmo variants follow the prompt format and return numerical answers, their predictions remain largely incorrect. The only models that surpass random accuracy are the Qwen2.5-VL models, with performance improving as the model size increases. Nonetheless, even these models fall short of robust comic understanding, highlighting the need for task-specific training.

\begin{table}[t]
\centering
\caption{Accuracy (in \%) of small (top) and large (bottom) models on \textit{ComicsPAP} tasks. Tasks are reported as: Sequence Filling, Character Coherence, Text Closure, Visual Closure, and Caption Relevance. In gray, results after finetuning.}
\label{tab:zero-shot}
\vspace{5mm}
\begin{tabular}{lc|ccccc|r}
\toprule
\textbf{Model}         & \textit{Size} & \textit{Sequence} & \textit{Character} & \textit{Visual} & ~~~\textit{Text}~~ &  \textit{Caption}~ & ~~\textit{Global} \\
\midrule
Random                 & 0     & 20.22 & 50.00 & 14.41 & 25.00 & 25.00 & 24.30 \\
\midrule
SmolVLM-256M           & 256M  & 1.52 & 2.10 & 1.00 & 1.16 & 1.91 & 1.47 \\
\rowcolor{lightgray!25} ~~~(fine-tuned)  & 256M  & 24.05 & \underline{58.74} & 13.33 & 24.32 & 25.19 & 25.77 \\
SmolVLM-500M           & 500M  & 0.38 & 2.10 & 1.00 & 0.38 & 2.29 & 1.14 \\
\rowcolor{lightgray!25} ~~~(fine-tuned) & 500M  & 18.70 & 50.34 & 14.00 & 24.32 & 24.42 & 23.65 \\
SmolVLM-2B             & 2B    & 4.20 & 7.69 & 3.67 & 5.58 & 4.96 & 4.98 \\
\rowcolor{lightgray!25} ~~~(fine-tuned)  & 2B    & 22.14 & 43.36 & 14.33 & 24.71 & 25.95 & 24.06 \\
PaliGemmav2-3B         & 3B    & 17.17 & 38.46 & 12.00 & 26.64 & 1.14 & 16.97 \\
Qwen2.5-VL-3B           & 3B    & 27.48 & 48.95 & 21.33 & 27.41 & 32.82 & 29.61 \\
\rowcolor{lightgray!25} ~~~(fine-tuned)  & 3B    & \underline{62.21} & \textbf{93.01} & \textbf{42.33} & \underline{63.71} & 35.49 & \underline{55.55} \\
\midrule
Molmo-O-7B             & 7B    & 17.18 & 39.86 & 12.33 & 22.39 & 27.48 & 21.94 \\
Molmo-D-7B             & 7B    & 16.03 & 9.09 & 15.67 & 17.76 & 25.57 & 17.54 \\
Qwen2.5-VL-7B             & 7B    & 30.53 & 54.55 & 22.00 & 37.45 & \underline{40.84} & 34.91 \\
\rowcolor{lightgray!25} ~~~(fine-tuned)   & 7B    & \textbf{69.08} & \textbf{93.01} & \underline{42.00} & \textbf{74.90} & \textbf{49.62} & \textbf{62.31} \\
Qwen2.5-VL-72B            & 72B   & 46.88 & 53.84 & 23.66 & 55.60 & 38.17 & 41.27 \\
\bottomrule
\end{tabular}
\end{table}

\subsection{Fine-tuning Experiments}
To mitigate the shortcomings observed in the zero-shot evaluations, we fine-tuned select models on our 100K training samples using supervised fine-tuning (SFT) with LoRA modules. Specifically, we fine-tuned the SmolVLM series in full precision and applied LoRA-based fine-tuning to the 3B and 7B Qwen2.5-VL models. The fine-tuning objective was twofold: (i) to teach the models to adhere to our task instructions by outputting responses in the format ``\texttt{answer: <choice number>}'', and (ii) to enable the models to internalize the three defined skills underlying comic strip understanding.

The fine-tuning is done on all five tasks simultaneously, sampling at random from each of the tasks during data loading. This means that within a single batch the model will see samples from different tasks. 
The training was performed using a constant learning rate of $2\text{e-}4$ with the AdamW optimizer. The LoRA configuration employed an $\alpha$ of $16$, a dropout rate of $0.05$, and a rank $r=8$. Table~\ref{tab:training_specs} summarizes the training configurations for our fine-tuning experiments across various models.

\begin{table}[t]
\centering
\caption{Training configurations for fine-tuning experiments. In the header, ``Size'' is number of parameters (expressed in Billion), ``GPU'' is the GPU type, ``MEM'' refers to GPU memory, ``BS'' is the effective training batch size computed as $\textit{Per Device BS}\times\textit{Num Devices}\times\textit{Gradient Accumulation Steps}$, ``\textit{steps}'' is the number of iterations done during training, and ``\textit{GPU hours}'' is computed as $\textit{Run Time}\times\textit{Num Devices}$.}
\label{tab:training_specs}
\vspace{5mm}
\begin{tabular}{l|cccccc}
\toprule
\textbf{Model}  & ~~\textit{Size}~~ & ~~\textit{GPU}~~ & ~~\textit{MEM}~~ & ~~~\textit{BS}~~~ & ~~\textit{steps}~~ & \textit{GPU hours} \\
\midrule
SmolVLM-256M    & 256M  & L40   & 48GB    & $128$  & $5,000$ & 19.2\\
SmolVLM-500M    & 500M  & L40  & 48GB    & $128$ & $5,000$ & 22.3\\
SmolVLM-2B      & 2B    & H100   & 64GB    & $128$ & $5,000$ & 62.75\\
\midrule
Qwen2.5-VL-3B   & 3B    & H100  & 64GB    & $128$ & $5,000$ & 98.69 \\
Qwen2.5-VL-7B   & 7B    & H100  & 64GB    & $128$ & $5,000$ & 127.36 \\
\bottomrule
\end{tabular}
\end{table}

As shown in Table~\ref{tab:zero-shot}, fine-tuning leads to substantial improvements. Although the fine-tuned SmolVLM models exhibit better performance compared to their zero-shot counterparts, their accuracies remain relatively low. In contrast, the fine-tuned Qwen2.5-VL models—particularly the 7B variant—demonstrate marked improvements across nearly all tasks, outperforming even the zero-shot Qwen2.5-VL-72B in global accuracy, despite one-tenth of the parameters. These results underscore both the challenges of comic panel understanding and the effectiveness of targeted fine-tuning.

\section{Conclusions}
We have presented \textit{ComicsPAP}, a novel benchmark designed to evaluate comic strip understanding in large multimodal models. By assembling a high-quality dataset through a combination of manual and automatic annotations, we have defined and assessed five key tasks—sequence filling, character coherence, visual closure, text closure, and caption relevance—that together probe the nuanced abilities required for comics comprehension. Our benchmarks thus serve a dual purpose: (i) to expose the inherent limitations of current LMMs in working with comic panel sequences, and (ii) to motivate the fine-tuning process using our automatically generated training set in order to improve these results. Our experimental results indicate that fine-tuning on our automatically annotated training set can boost model performance by more than $30\%$ over their zero-shot baselines, surpassing $10x$ larger models.

\paragraph{\textbf{Open Research Directions.}} Despite promising results, several avenues remain open for future investigation. First, our fine-tuning strategy used uniform sampling across all skills, but adaptive sampling strategies accounting for task difficulty may improve performance. Also, instead of a single model trained of multiple tasks, it could be interesting to see performances of multiple task-specific fine-tuned models. Second, prompts in our experiments were generic; designing task-specific prompts could help models better utilize contextual cues. Third, while our study focused on foundational LMMs, emerging specialized models—such as Magi~\cite{sachdeva_manga_2024}, which provides graph-based representations of comic elements—could significantly enhance performance if integrated with Graph Neural Networks (GNNs). Fourth, combining multiple smaller, task-specific models into an ensemble might yield superior results with fewer parameters than larger monolithic architectures. Finally, we designed the tasks to leverage automatic annotations with some human refinement. However, we believe the quality of captions and OCR outputs can significantly influence model performance; thus, comparing earlier-generation extraction methods with newer or manually crafted alternatives remains an important direction for future work.
Exploring these directions promises to further advance multimodal comic comprehension.

\ifblind

\else
\section*{Acknowledgements}
This paper has been supported by the Consolidated Research Group 2021 SGR 01559 from the Research and University Department of the Catalan Government, and by project PID2023-146426NB-100 funded by MCIU/AEI/10.13039/ 501100011033 and FEDER, UE. With the support of the FI SDUR predoctoral grant program from the Department of Research and Universities of the Generalitat de Catalunya and co-financing by the European Social Fund Plus (2024FISDU\_00095). \newline

We acknowledge EuroHPC Joint Undertaking for awarding the project ID EHPC-AI-2024A02-057 access to MareNostrum5 at BSC, Spain.

\fi

%
% ---- Bibliography ----
%
\bibliographystyle{splncs04}
\bibliography{references}

\end{document}